\documentclass{article}

\usepackage{color}

\usepackage[final,nonatbib]{tackling_climate_workshop_style}

\usepackage[utf8]{inputenc} 
\usepackage[T1]{fontenc}    
\usepackage{hyperref}       
\usepackage{url}            
\usepackage{booktabs}       
\usepackage{amsfonts}       
\usepackage{nicefrac}       
\usepackage{microtype}      
\usepackage{lipsum}
\usepackage[pdftex]{graphicx}
\usepackage{subfig}
\usepackage{wrapfig}
\usepackage{authblk}
\graphicspath{ {./images/} }

\newcommand*\samethanks[1][\value{footnote}]{\footnotemark[#1]}

\title{SunCast: Solar Irradiance Nowcasting \\from Geosynchronous Satellite Data}

\author[{ } { }1]{Dhileeban Kumaresan\thanks{Equal Contribution}}
\author[{ } { }1]{Richard Wang\samethanks}
\author[{ } { }1]{Ernesto Martinez\samethanks}
\author[{ } { }1]{Richard Cziva\samethanks}
\author[1]{\authorcr Alberto Todeschini}
\author[2]{Colorado J Reed}
\author[1]{Hossein Vahabi}
\affil[1]{School of Information, University of California, Berkeley, USA}
\affil[2]{EECS / BAIR, University of California, Berkeley, USA}
\affil[ ]{\texttt {\{dkumares,cweiwang,ernestomartinez,rcziva,todeschini,cjrd,puyavahabi\}@berkeley.edu}}

\begin{document}
\maketitle
\begin{abstract}

When cloud layers cover photovoltaic (PV) panels, the amount of power the panels produce fluctuates rapidly. Therefore, to maintain enough energy on a power grid to match demand, utilities companies rely on reserve power sources that typically come from fossil fuels and therefore pollute the environment. Accurate short-term PV power prediction enables operators to maximize the amount of power obtained from PV panels and safely reduce the reserve energy needed from fossil fuel sources. While several studies have developed machine learning models to predict solar irradiance at specific PV generation facilities, little work has been done to model short-term solar irradiance on a global scale. Furthermore, models that have been developed are proprietary and have architectures that are not publicly available or rely on  computationally demanding Numerical Weather Prediction (NWP) models. Here, we propose a Convolutional Long Short-Term Memory Network model that treats solar nowcasting as a next frame prediction problem, is more efficient than NWP models and has a straightforward, reproducible architecture. Our models can predict solar irradiance for entire North America for up to 3 hours in under 60 seconds on a single machine without a GPU and has a RMSE of 120 $W/m^2$ when evaluated on 2 months of data.

\end{abstract}

\section{Introduction}

Solar photovoltaic (PV) power generation has gained significant popularity and attention in the past few decades. According to studies, it is expected to account for almost 80\% of the increase in renewable energy generation through 2050 \cite{eia2021}. Moreover, solar power is required to reduce the $\sim$25\% of global greenhouse gas emissions that stem from the energy sector each year \cite{owusu2016,elzinga2015energy}. Despite its popularity, environmental-friendliness and the vast amount of energy available, solar power is associated with intermittency and uncertainty -- its output can substantially rise or fall  instantaneously based on the cloud layers between the sun and the solar panels. 

To avoid power blackouts and damage of equipment, electricity grids need to remain balanced within a narrow range at all times. This is typically accomplished by utilizing backup generators~\cite{perkins2018techno}, and sometimes using battery reserves~\cite{ghennam2015advanced} and demand-response scheduling~\cite{wang2012short}. However, these solutions have their own limitations. As for the most popular solution, backup generators run on fossil fuels and their power ascend and decline rate is restricted by the unit ramp rate which results in difficulties to meet incremental power generation needs~\cite{yuan2014improved}.  Large-scale battery technologies on the other hand are expensive and difficult to realize. Lastly, demand-response scheduling is challenging due to the lack of information on temporal consumption and power available. Near-term accurate forecasting of solar energy production may allow a utility to store energy in advance or avoid spinning reserves thus improving cost-effectiveness and security \cite{haupt2019, wu2015, antonanzas2016, kaur2016}. 

Recently, solar power prediction has been widely studied using ground data and satellite imaging. For example, Hu et al. uses ground-based images to predict production at a solar plant in Hangzhou, China~\cite{hu2018}. There has also been much research aimed at predicting PV power over a geographical range, such as Mathe et al.~\cite{mathe2019}, where they trained a deep learning model on satellite images in order to predict total daily PV output for the entire nation of Germany. Numerical weather prediction (NWP) models (such as one used by Verbois et al.~\cite{verbois2018solar}) are often used for solar irradiation prediction, which involve extensive computational overhead and are more suitable for predicting long-term horizons (e.g. days). Furthermore, utility-scale solar farms invest in developing their own models to predict power production but the distributed nature of solar installations with millions of small sites encourages automated and efficient prediction models for a wide area.

In this paper, we predict solar irradiance using a Convolutional LSTM model architecture for the entire North America. We use satellite images of irradiance and treat this as a next-frame prediction problem; this allows for the development of a model that is simple, computationally efficient, and reproducible. Since each pixel in the predicted frame represents a geographic area, predictions can be made at both the local and national scales. Our models predict solar irradiance for North America for up to 3 hours in under 60 seconds on a single machine without a GPU and has a RMSE of 120 $W/m^2$ when evaluated on 2 months of data.

\section{Data and Methodology}
\label{sec:headings}
A standard approach to predict future solar irradiance consists of using weather features to compute and predict irradiance. Many factors must be considered to determine solar irradiance such as cloud type, cloud depth, aerosol type, aerosol depth, ozone, wind speed, wind direction, ground location and solar zenith angle. This information is needed for every prediction location, which makes traditional numerical modeling processes very complex and computationally intensive. Instead of approaching this as an numerical weather problem, we approach it as a computer vision problem using the GOES-16 data as described below.

\subsection{Data (GOES-16)}

Geostationary Operational Environmental Satellite 16 (GOES-16), operated by NASA, is a weather satellite that continuously provides downward shortwave radiation at the surface (DSR) data. DSR is the total amount of shortwave radiation (both direct and diffuse) that reaches the Earth’s surface. The solar energy industry needs estimates of DSR for both real-time and short-term forecasts for building energy usage modeling and optimization \cite{dsr2018}.
This data is updated hourly and automatically processed by our pipeline from \emph{Registry of Open Data on AWS}\footnote{https://registry.opendata.aws/noaa-goes/}.

\subsection{Methodology}

Our goal  is to predict DSR values for North America for 3 hours into the future. Three DSR images of size 166 x 394 belonging to hours \emph{t-2, t-1 and t} are used as input to the model to predict DSR images for \emph{t+1, t+2 and t+3}. Using 3 input images, provide the model with not only the historical DSR values, but also how it moved in the last 3 hours. One-hot encoding of month and hour of the day are also embedded in these input images to account for sun's position \cite{sunangle}.

\subsubsection{Linear Regression}
Linear regression was selected as a simple baseline. Last 3 hours of DSR are used as our input to predict next hour's DSR. Every DSR image consist of 65,404 pixels which means every input has a dimension of 196,212 and prediction has a dimension of 65,404. A linear model trained on this input/output configuration will be very large, and it would undermine our goal of developing a simple and efficient approach. To increase efficiency, we scaled down to make pixel-wise predictions for the linear model. The model output is a single pixel and the input is 40 by 40 image sample from the DSR images from previous 3 hours. The result shows that the model assigned fixed weights to prior hour pixels. Hence, spatio-temporal movements of the DSR are not learned. Most other models not leveraging LSTM will have similar problem where fixed weights are assigned to pixels rather than learning the pattern of movements.

\begin{figure}[t]
    \centering
    \subfloat[Downward Shortwave Radiation (DSR) at the surface from the GOES-16 Satellite. Values are in $W/m^2$.]{{\includegraphics[width=4cm]{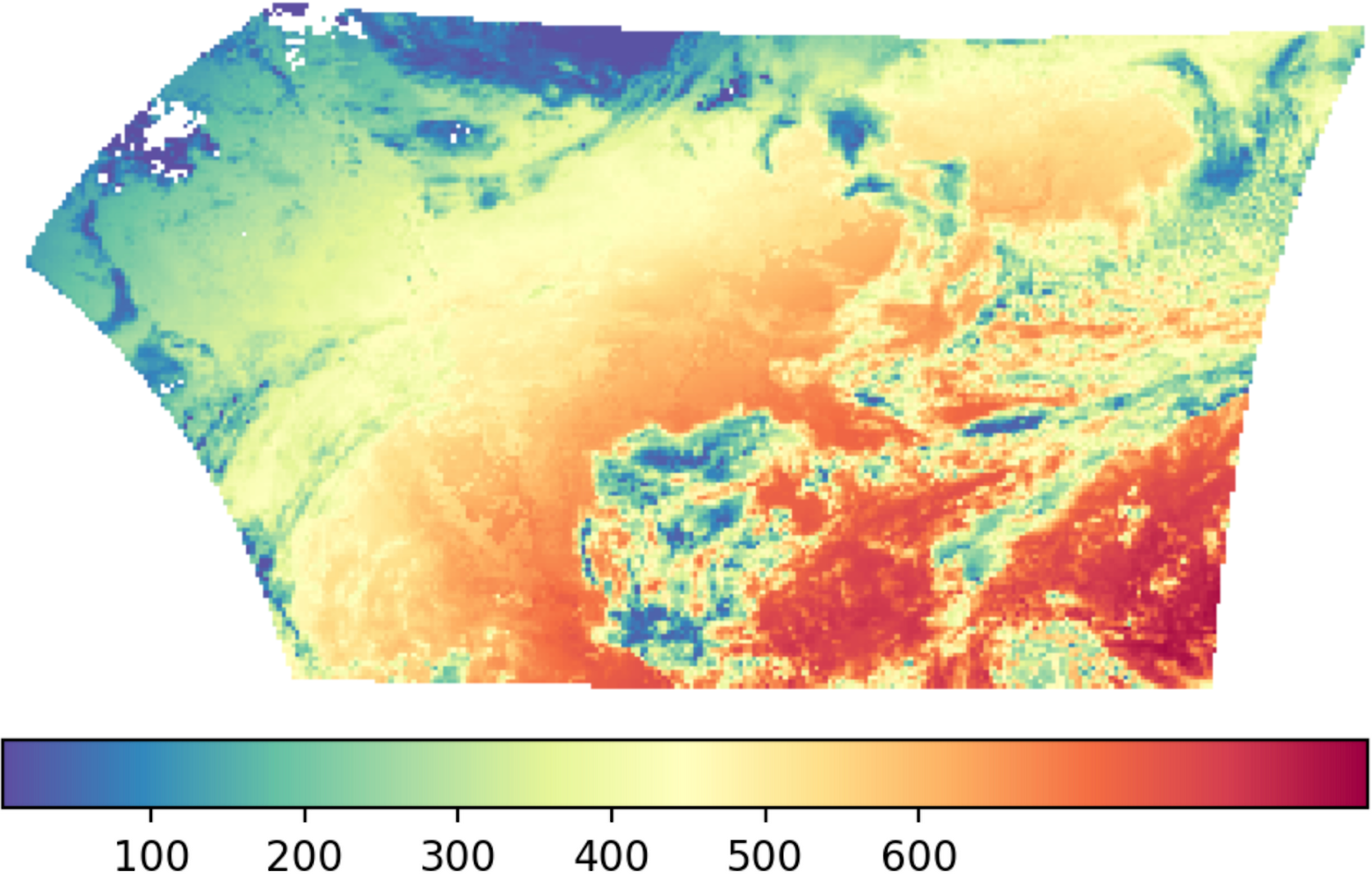} }}%
    \qquad
    \subfloat[ConvLSTM Architecture. Three pre-processed DSR images are passed into 3 layers of ConvLSTM layers each separated by Batch Normalization layer and finally through a Conv3D layer to predict 3 future images.]{{\includegraphics[width=7cm]{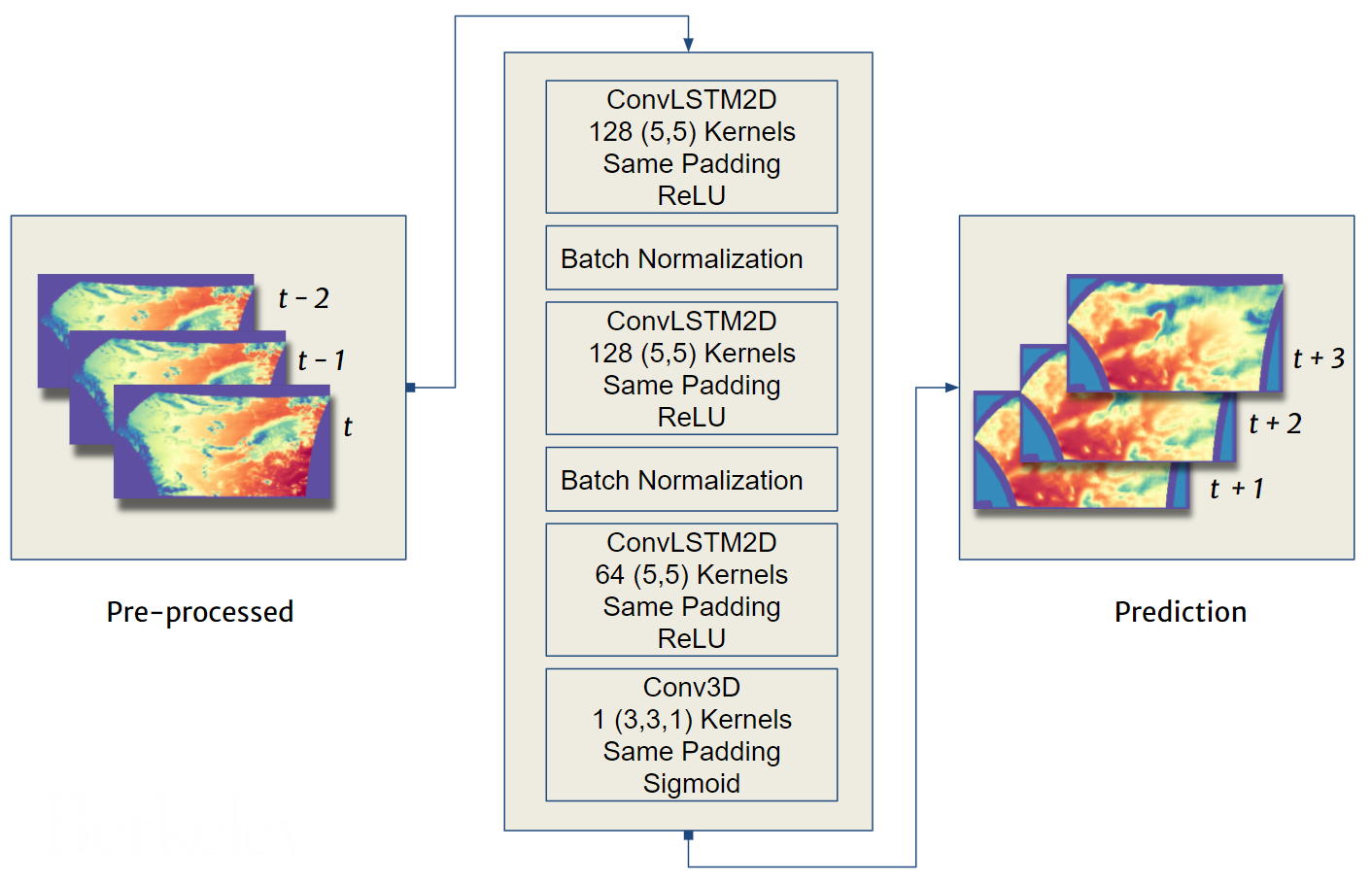} }}%
    \label{fig:example}%
\end{figure}

\subsubsection{Convolutional Long Short-Term Memory Network (ConvLSTM)}

The goal of the model is to precisely predict the downward shortwave radiation for a relatively short period of time, for e.g. 0-3 hours. For the sake of computational efficiency and to predict the DSR for a wider area, we treat this problem as a spatiotemporal sequence prediction problem. Shi et al\cite{xingjian2015convolutional} formulated precipitation now-casting as a spatiotemporal sequence forecasting problem that can be solved under the general sequence-to-sequence learning framework proposed in \cite{sutskever2014}. In order to model the spatiotemporal relationships, they extend the idea of FullyConnected-LSTM to ConvLSTM which has convolutional structures in both the input-to-state and state-to-state transitions. By stacking multiple ConvLSTM layers and forming an encoding-forecasting structure, we build an end-to-end trainable model for nowcasting downward shortwave radiation.

Our network is comprised of 2 ConvLSTM layers with 128 kernels of size 5x5 separated by a batch normalization layer followed by another ConvLSTM layer with 64 kernels of size (5,5). All the ConvLSTM layers use same padding and ReLU activation. This is followed by a regular Conv3D layer to produce output of the desired shape. This uses same padding and a sigmoid activation to squash the values between 0 and 1. We trained our model using MSE loss.

Our experiments indicate that training different models for predicting each of the 3 hours were more accurate than one model predicting all three hours' DSR images. The model trained to predict first hour DSR has the output sequence \emph{t-1, t, t+1}, second hour model has output sequence \emph{t, t+1, t+2} and third hour has \emph{t+1, t+2, t+3}. The last image in each output sequence is the prediction.

\paragraph{Cell Variations - }
To see if we can improve the results from ConvLSTM model, we experimented modifying the ConvLSTM cell by introducing a peephole connection and let the gate layers inspect the cell state of the same memory block \cite{ger2000}. We call this a ConvLSTMPeephole cell. In another experiment, we combined the input and forget gates into an update gate to see if it can deliver results similar to ConvLSTM model by improving computational efficiency \cite{cho2014,chung2014empirical}. We call this a ConvGRU cell. In both these cell variations, we replaced the Hadamard product with the convolution operation to match with the ConvLSTM cell.

\section{Results and Discussion}\label{sec:headings}

\subsection{Error Evaluation}

 Table~\ref{tab:performance}  provides a summary of the RMSE values for our predictions made for the two-month test set. These results are broken down by prediction hour as well as the range of the true DSR value. Exploring any potential differences in performance based on DSR ranges may be useful for demonstrating that our model is effective in a variety of weather/climatic conditions. From the results in Table~\ref{tab:performance}, we see that the ConvLSTMPeephole model outperforms the other models for the first and third hour predictions, while the unaltered ConvLSTM model performed slightly better for the second hour prediction.

\begin{table}[t]
  \centering
  \scalebox{0.7}{
      \begin{tabular}{lllll}
        \toprule
        \multicolumn{1}{c}{\textbf{First Hour Prediction}}
        \\
        \cmidrule(r){1-2}
        Model               & Overall        & Low DSR (0-300) & Medium DSR (300-600) & High DSR (600+) \\
        \midrule
        ConvLSTM            & 61.4          & 56.3           & 74.6          & 67.5          \\
        ConvLSTMPeephole & \textbf{60.2} & \textbf{55.2} & \textbf{73.1} & \textbf{66.1} \\
        ConvGRU                 & 97.2          & 77.3          & 152.1         & 107.1         \\
        \midrule
        \multicolumn{1}{c}{\textbf{Second Hour Prediction}}
        \\
        \cmidrule(r){1-2}
        Model               & Overall        & Low DSR (0-300) & Medium DSR (300-600) & High DSR (600+) \\
        \midrule
        ConvLSTM            & \textbf{85.7} & \textbf{76.3} & \textbf{117.8} & \textbf{88.1} \\
        ConvLSTMPeephole & 86.5          & 77.9          & 119.1          & 86.3          \\
        ConvGRU                 & 126.9         & 96.9          & 209.5          & 137.9         \\
        \midrule
        \multicolumn{1}{c}{\textbf{Third Hour Prediction}}
        \\
        \cmidrule(r){1-2}
        Model               & Overall         & Low DSR (0-300) & Medium DSR (300-600) & High DSR (600+) \\
        \midrule
        ConvLSTM            & 123.8          & 102.3         & \textbf{184.6} & 136.2          \\
        ConvLSTMPeephole & \textbf{120.6} & \textbf{88.2} & 205.9          & \textbf{132.5} \\
        ConvGRU                 & 159.2          & 93.5          & 267.7          & 223.9          \\
        \bottomrule
      \end{tabular}
    }
     \caption{Model Performance on the 2-month test set(RMSE, $W/m^2$).}
  \label{tab:performance}
\end{table}

\paragraph{Baseline Comparison}

Test evaluation figures from Table~\ref{tab:performance} are based on the full DSR images. However, due to the computational challenges with the linear regression, the baseline was tested on a subsection of the DSR images encompassing the state of California. Predictions made by the baseline for this subsection were compared with our ConvLSTM model. The Linear Regression and ConvLSTM models had a RMSE of 104.6 and 71.4 $W/m^2$ respectively.

\subsection{Comparison to Numerical Weather Prediction}

We compared predictions made by the ConvLSTMPeephole model to High Resolution Rapid Refresh (HRRR), a numerical weather prediction model that is operated by the U.S federal government~\cite{benjamin2016} and requires more than 10 TB of memory to run~\cite {hrrr2020}. We examined predictions at 22 sites (17 large solar farm sites and 5 urban centers) between the hours of 10:00AM to 3:00PM PST for four weeks of the test set. The results are summarized in Table~\ref{tab:hrrr}. Predictions made by our model resulted in a 13.08\% decrease in RMSE. These results demonstrate that by approaching DSR prediction as a next frame prediction problem, we can achieve results that are comparable to NWP models while avoiding the computational overhead.

\begin{table}[h]
  \centering
  \scalebox{0.7}{
      \begin{tabular}{lll}
        \toprule
        Grouping             & HRRR RMSE & Model RMSE \\
        \midrule
        Overall              & 124.9    & 108.6      \\
        Low DSR (0-300)      & 165.3    & 135.3     \\
        Medium DSR (300-600) & 170.7    & 131.7     \\
        High DSR (600+)      & 103.5    & 98.3      \\
        \bottomrule
      \end{tabular}
  }
  \caption{ConvLSTM vs HRRR performance for predictions made for 22 locations between 10:00AM-3:00PM PST four four weeks of the test set (RMSE, $W/m^2$).}
  \label{tab:hrrr}
\end{table}

\section{Conclusion}
\label{sec:headings}

In this paper, we looked at short term solar irradiance prediction as a next-frame-prediction problem and is advantageous in several ways over NWP models. The data used for training and testing is made publicly available by NOAA, and can be used as inputs to the model with minor pre-processing. Once trained, running our models require only a single virtual computer without a GPU. Inference time is under 60 seconds, meaning predictions can be delivered to the end user with enough time to make use of the information. We have made predictions available via our REST API, which is updated hourly to provide predictions in real-time. Future work may involve including more number of input images, adding channels (e.g. infrared, near-infrared and visible) to these input images to see how this may improve performance, as well as adapting self-supervised and domain adaptation methods to this work~\cite{Reed_2022_WACV,xiao2021region,yue2021multisource}. In addition, NOAA operates geostationary satellites that collect similar spectral data all across the globe. Training our model on data from other regions around the world would provide insight about the effectiveness of our approach on a global scale.

\bibliographystyle{unsrt}  
\bibliography{references}

\end{document}